\documentclass{l4dc2025}

\usepackage{lipsum}
\usepackage{amsmath}
\usepackage{amsfonts}
\usepackage{mathtools}          
\usepackage{mathrsfs} 
\usepackage{multirow}
\usepackage{adjustbox}
\usepackage{booktabs}
\usepackage{colortbl}
\usepackage{wasysym}
\usepackage{pifont}
\usepackage{algorithm}
\usepackage{algorithmic}
\usepackage{wrapfig}
\usepackage{graphicx}
\usepackage{caption}

\newcommand{\unsat}{\texttt{UNSAT}}

\newcommand{\realizability}{\texttt{realizability}}

\newcommand{\propershield}{\texttt{Proper Shield}}
\newcommand{\cmark}{\textcolor{green}{\ding{51}}}
\newcommand{\xmark}{\textcolor{red}{\ding{55}}}

\title[Continuous Safety Shield for RL]{Realizable Continuous-Space Shields for Safe Reinforcement Learning}
\usepackage{times}

\coltauthor{\Name{Kyungmin Kim}\textsuperscript{\rm 1,}\footnote{These authors contributed equally to the work.}, 
\Name{Davide Corsi}\textsuperscript{\rm 1,\rm \tiny{*}}, 
\Name{Andoni Rodríguez}\textsuperscript{\rm 2,\rm 3,\rm \tiny{*}}, \\
\Name{JB Lanier}\textsuperscript{\rm 1}, 
\Name{Benjami Parellada}\textsuperscript{\rm 4}, 
\Name{Pierre Baldi}\textsuperscript{\rm 1},
\Name{César Sánchez}\textsuperscript{\rm 2}, 
\Name{Roy Fox}\textsuperscript{\rm 1} \\
\addr 
    \textsuperscript{\rm 1}University of California, Irvine, USA \\
    \textsuperscript{\rm 2}IMDEA Software Institute, Madrid, Spain\\
    \textsuperscript{\rm 3}Universidad Politécnica de Madrid, Madrid, Spain\\
    \textsuperscript{\rm 4}Universitat Politècnica de Catalunya, Barcelona, Spain}

\begin{document}

\maketitle

\begin{abstract}%
While Deep Reinforcement Learning (DRL) has achieved remarkable success across various domains, it remains vulnerable to occasional catastrophic failures without additional safeguards. An effective solution to prevent these failures is to use a shield that validates and adjusts the agent's actions to ensure compliance with a provided set of safety specifications. For real-world robotic domains, it is essential to define safety specifications over continuous state and action spaces to accurately account for system dynamics and compute new actions that minimally deviate from the agent’s original decision. In this paper, we present the first shielding approach specifically designed to ensure the satisfaction of safety requirements in continuous state and action spaces, making it suitable for practical robotic applications.
Our method builds upon \texttt{realizability}, an essential property that confirms the shield will \textbf{always} be able to generate a safe action for \textit{any} state in the environment. We formally prove that \realizability{} can be verified for stateful shields, enabling the incorporation of non-Markovian safety requirements, such as loop avoidance. Finally, we demonstrate the effectiveness of our approach in ensuring safety without compromising the policy’s success rate by applying it to a navigation problem and a multi-agent particle environment
\end{abstract}



\section{Introduction}
\label{sec:introduction} 

Deep Reinforcement Learning (DRL) has achieved impressive results in a wide range of fields, from mastering complex games like Go \citep{silver2016mastering} and Dota 2 \citep{berner2019dota} to real-world applications in healthcare \citep{pore2021safe}, autonomous driving \citep{tai2017virtual}, and robotics \citep{aractingi2023controlling}. However, even advanced DRL algorithms \citep{schulman2017proximal} face considerable challenges when analyzed on specific corner cases, where they persist in demonstrating a proclivity to commit critical mistakes \citep{corsi2024analyzing, szegedy2013intriguing}. Such limitations present a threat to the reliability of DRL systems, particularly when deployed in safety-critical applications, where even a single failure can have potentially catastrophic consequences \citep{srinivasan2020learning, marvi2021safe, katz2019marabou, corsi2021formal}. 

Traditional techniques to address safety concerns aim to embed this aspect as part of the learning process; some examples include sophisticated reward shaping techniques \citep{tessler2018reward}, constrained reinforcement learning \citep{achiam2017constrained, ray2019benchmarking}, and adversarial training \citep{pinto2017robust}. While these approaches can significantly enhance the overall reliability of the policy, their guarantees are typically empirical and their benefits are provided only in expectation \citep{srinivasan2020learning, he2023autocost}. Although these may be sufficient for many problems, they cannot guarantee the safety of a system, limiting their applicability in highly safety-critical contexts.


\begin{figure}[t]
\centering
\includegraphics[width=0.7\linewidth]{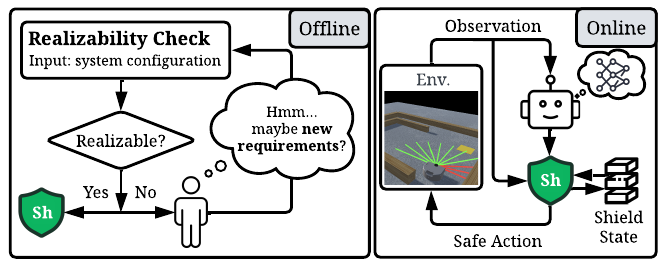} 
\caption{In the offline process, the realizability check verifies whether the system is realizable, taking the system configuration as input, i.e., the dynamics of the environment, the safety requirements, and the input domain. If not realizable, the system needs to be changed. When proved to be realizable (i.e., a \propershield{}), the shield can be safely deployed in the environment for the online process.}
\label{fig:intro:main}
\vspace{-15pt}
\end{figure}

To face this limitation, a promising family of approaches provides formal safety guarantees through the adoption of an external component, commonly referred to as a shield \citep{garcia2015comprehensive, corsi2024verification}.
A shield acts as a protective wrapper over the agent to ensure that its actions remain within safe boundaries, effectively preventing it from making dangerous or undesired decisions \citep{alshiekh2018safe}. However, most shielding techniques in the literature face a fundamental challenge: there are states where no action satisfies all safety criteria simultaneously.
This issue is particularly critical in robotics. When an agent encounters a scenario where no safe action is available from the shield, it is left uncertain about how to proceed, as a default safe action is not always feasible in complex and dynamic environments. 
Therefore, it is essential to develop a shield that can always provide an action satisfying the given specifications for \textit{any} state of the system, which we define as a \propershield.

In the logic community, ensuring this property is equivalent to guaranteeing the realizability of the shield.
\citet{alshiekh2018safe} proposed a method to design a proper shield using Linear Temporal Logic (LTL). Their approach uses the synthesis of the shield as a formal tool to automatically guarantee the realizability of the system.
However, this method is limited to discrete states and actions, as the LTL synthesizer is not designed to handle continuous spaces. This constraint is overly restrictive for real-world reinforcement learning applications such as robotics where the state and action spaces are often continuous, high dimensional and nonlinear. 
More in general, despite its necessity and importance, little has been explored regarding continuous-space proper shields in robotics, Indeed, compared to discrete spaces, ensuring a proper shield in continuous spaces is significantly more difficult and remains an open challenge. 

In this work, we extend a proper shield to continuous spaces, addressing a gap in existing methods. 
Our approach builds on the theoretical foundations provided by \citet{rodriguez2023boolean}, which solve the realizability problem for some fragments of \textbf{LTLt}, an extension of LTL that allows to include real values in the specification.
While their work focuses purely on logical formulations with no direct connection to sequential decision-making, we adapt these ideas to develop a practical framework for generating a proper shield for continuous control tasks.
As shown in Fig.~\ref{fig:intro:main}, realizability checks are part of an offline procedure conducted prior to deployment. Crucially, working in continuous spaces allows us to encode a more accurate, model-based representation of the physical world within the specifications. 
Additionally, working with real values allows us to integrate optimization techniques to ensure that the shield not only returns safe actions but also provides decisions closely aligned with those proposed by the agent. This ensures the agent’s behavior remains both effective and optimal while adhering to strict safety constraints.

Additionally, we formally prove that realizability can be verified for shields with non-Markovian safety requirements.
While the realizability of continuous non-Markovian requirements is typically undecidable, we address this challenge by introducing a so-called \textit{anticipation} fragment of LTLt that eliminates operators requiring infinite memory, assuming deterministic dynamics.
This fragment is expressive enough to capture relevant safety requirements under assumptions that are common in practical scenarios.
For example, in our case study, we encode a rule to avoid loops within a specified time window (e.g., do not visit the same region for $i$ timesteps) to enhance success rates. 

We demonstrate the effectiveness of our shielding approach in a mapless navigation domain, with additional experiments in a particle world multi-agent environment.
In the mapless navigation domain, the shield effectively protects the agent from obstacles in an unknown environment, highlighting the shield's ability to enforce safety without relying on pre-defined maps or global knowledge.
We guarantee the realizability of safety requirements, ensuring that the agent has always at least one safe action available. Furthermore, with the integration of optimization techniques and the non-Markovian requirements we achieve the safety of the system with a minimal impact on the policy performance.
In the particle-world multi-agent environment, our shield manages interactions among multiple agents in a shared, continuous space. 

In summary, the contributions of this paper are as follows: (1) We introduce continuous-space proper shields for safe control. A detailed comparison with alternative shielding techniques is provided in Tab.~\ref{tab:introduction:summary} in the Appendix. (2) We propose a shield that enables formal realizability proofs for non-Markovian requirements by introducing the ‘anticipation fragment’ of LTLt, a decidable subset designed for stateful shields. This fragment effectively captures safety requirements under reasonable assumptions for practical use.
(3) We demonstrate the merits of our method on robotic applications, including a mapless navigation domain and a particle-world multi-agent environment.

\section{Preliminaries}
\label{sec:preeliminaries}

We model interaction with an environment with a \textbf{Partially Observable Markov Decision Process (POMDP)}, defined by a tuple $(X, A, O, P, R, \Omega, \gamma)$, with a state space of $X$, action space $A$, and observation space $O$. $P: X \times A \to \Delta(X)$ represents the state transition function. $R: X \times A \to \mathbb{R}$ is the reward function. $\Omega: X \to \Delta(O)$ represents the observation function, and $\gamma \in [0, 1]$ is the discount factor. Our objective is to learn a policy $\pi: O \to \Delta(A)$ that maximizes the expected sum of discounted rewards over time. This is defined as $ \mathbb{E} \left[ \sum_{t=0}^{T} \gamma^t R(x_t, a_t) \right] $, where $x_t$, $a_t$, and $o_t$ are the state, action, and observation at time $t$ until the episode horizon $T$.

\subsection{Concepts in Temporal Logic and Reactive Systems}

\paragraph{LTL.}
Linear Temporal Logic (LTL) is a formal system for reasoning about the behavior of discrete-time systems over an infinite timeline. Concretely, it is a modal logic that extends propositional logic with temporal operators that allow the expression of properties about the future evolution of the system.
Some key temporal operators in LTL include: (1) $\ocircle$ (next): The property holds in the next timstep; and (2) $\square$ (always): The property holds at all future timsteps.
%
Using these operators, LTL formulas can express a wide range of temporal properties, such as safety (``something bad never happens").
%
For instance, while in classic propositional logic we can express $v^1 \rightarrow v^2$, in LTL we can also express $\square(v^1 \rightarrow \ocircle v^2)$, which means that at every step, if $v^1$ holds, then $v^2$ must hold in the next timestep.

\paragraph{Satisfiability and Realizability.}
Given an LTL formula $\varphi$, the satisfiability problem determines if there is any possible execution 
that satisfies the specified temporal properties; i.e., 
we say $\varphi$ is satisfiable if there is a possible assignment of the variables in $\varphi$ such that $\varphi$ is satisfied.
More formally, given variables $v = \{v^0, v^1, ...\}$ in $\varphi$, if $\exists v$ s.t. $\varphi(v)$ holds, then $\varphi$ is satisfiable. 
%
A \textbf{reactive system} is a type of system that continuously interacts with its environment by responding to inputs and adapting its behavior accordingly.
To ensure such a system can meet the requirements under all possible conditions, we need a property stronger than satisfiability: \realizability.
In \realizability, the variables of $\varphi$ are divided into an uncontrollable player (i.e., the inputs provided by the environment) and a controllable player (i.e., the outputs provided by the system). 
Then, a formula $\varphi$ is realizable if for \textit{all} possible valuations of the input variables, the output variables can be assigned so that the $\varphi$ is not violated. 
\paragraph{Realizability for Continuous Domains.}
LTL is a powerful tool for reasoning about temporal properties. However, it lacks the expressiveness needed to handle continuous values, which are essential for realistic robotics applications to accurately encode environment dynamics in continuous spaces.
To address this, we use Linear Temporal Logic modulo theories (LTLt), an extension of LTL that allows formulas to include variables from a first-order theory $\mathcal{T}$.
%
While this has a precise meaning in formal verification \footnote{We invite the reader to read the supplementary material (Appendix~\ref{sec:firstOrder}) for further insights.}, for simplicity in this paper, it means that LTLt enables specifying more complex properties involving both the temporal behavior of a system and real values data.
A recent development by \citet{rodriguez2024realizability} solved the problem of realizability for certain fragments of LTLt.
This makes it feasible to create realizable continuous-space shields using LTLt to ensure such expressive properties; for example, $(v^1>2.5) \rightarrow  \ocircle(v^2>v^1)$.
%
Note that another powerful operator in LTLt is $\lhd v$, which, given $v$ allows access the value of $v$ in the previous timestep. However, this expressivity comes at the expense of the LTLt not being undecidable anymore (i.e., realizability checking procedures are not guaranteed to terminate).

\section{Realizable Continuous Shields with Non-Markovian Requirements}
\label{sec:method}

A deep reinforcement learning shield is an external component that works on top of a policy to ensure compliance with a set of safety requirements $\varphi$.  
A \propershield{} is a special case of a shield that can provide safe alternatives to unsafe actions for \textit{all} states in the system.

In this paper, to address the complexity of real-world robotic problems, our shield is designed to be \textit{stateful}, enabling it to handle non-Markovian requirements. 
The shield maintains a state $h \in H$ in which, 
by construction, $h \in H$ represents only safe states (i.e., states where $\varphi$ has not been violated), as any violation in the past would contradict the safety specifications. 
Formally, given an action $a \in A$, an observation $o \in O$, a shield state $h \in H$, and a set of safety specifications $\varphi$, we define a \propershield{} as follows:

\begin{definition}[\propershield{}]
\label{def:propershield} 
{\ \\}
\fbox{\begin{minipage}{1.0\textwidth}
A \propershield{} is a function $\zeta: A \times O \times H \to A$, such that $\forall$ \( o \in O \) and \( h \in H \), $\varphi(\zeta(a, o, h),o, h)$ holds.
\end{minipage}}
\end{definition}

Intuitively, a \propershield{} is a function $\zeta$ designed to always return an action $a$ such that $\varphi$ holds when starting from the initial state. The shield $\zeta$ is combined with an external agent $\pi$ (in our case, an RL agent) to ensure that the composition $\pi \cdot \zeta$ never violates $\varphi$. At each time step, given the shield's history $h$: (1) $\pi$ receives observation inputs $o$ from the environment and produces an action $a$; (2) we check whether $\varphi(a, o, h)$ holds; and (3) if $\varphi(a, o, h)$ holds, the shield does not intervene, but if there is a violation, it overrides $a$ with a corrected action $\hat{a} = \zeta (a,o,h)$ such that  $\varphi(\hat{a},o,h)$ now holds for that step, as described in Fig.~\ref{fig:intro:main} (right).

Ensuring that $\varphi$ is realizable is essential, as unrealizability could lead to situations where no safe action is possible for a given state. In such cases, computing $\hat{a}$ is not possible because the formula is unsatisfiable (\unsat{}). In our framework (see Fig.~\ref{fig:intro:main}, left), we check the realizability of $\varphi$ as part of an offline process prior to deployment, treating the observation $o$ and shield state $h$ as uncontrollable variables, while considering the action $a$ as a controllable variable. 

\citet{alshiekh2018safe} showed that a proper shield can be constructed using LTL, but their approach is limited to discrete state and action spaces.
This restricts its applicability to robotics tasks and fails to capture accurate environment dynamics.
In contrast, our shield leverages the more expressive specification language, LTLt, enabling the construction of a continuous-space proper shield that incorporates accurate dynamics directly into the specification. Moreover, since the continuous domain is a continuous space, we can minimize a distance metric between the correction $\hat{a}$ and the original candidate output $a$. This allows us to optimize the shield's safe-action correction by returning $\hat{a}$ such that it is the closest safe value to the agent's proposed action $a$.

\subsection{Non-Markovian Encoding in LTLt} \label{subsec:nonMarkov}

By exploiting recent advancements in LTLt synthesis as discussed in the previous sections, it becomes possible to encode Markovian properties that can be synthesized to create a proper shield:

\begin{example} \label{ex:markov}
    Consider a simple agent that moves along a one-dimensional line, where its state is represented by a single variable $x$. The action space consists of a single action $a$, which specifies both the direction and magnitude of the step. The dynamics are given by $x_{t+1} = x_{t} + a_{t}$. Suppose we wish to enforce a simple safety requirement that ensures the agent never moves to a position lower than zero. This requirement can be encoded as follows:
    \begin{equation*}
        [x < 1] \rightarrow \neg[a<-1]
    \end{equation*}
\end{example}

\noindent Note that this requirement is highly simplified; the challenge for the solver arises from the exponential growth in the number and complexity of such requirements. Nevertheless, properties of this form are often sufficient to encode essential safety requirements, such as collision avoidance. However, for real-world problems, we often need to incorporate requirements that extend beyond simple input-output relationships, involving multiple steps in the environment and sequences of the agent’s decisions. In such cases, it is necessary to encode non-Markovian requirements (or multistep properties) in a decidable fragment of LTLt, that guarantees an automatic realizability check. To better explain our approach we rely on a running example:

\begin{example} \label{ex:non-markov}
Let us revisit the agent from Example \ref{ex:markov}. We may now want to ensure that, once the agent visits a specific region of the state space, it will never return to the same position within a finite time horizon $i$. Specifically, we subdivide the state space in a finite number of intervals $r\in R$, where $r=[r_l, r_u]$. This requirement requires a non-Markovian constraint as it involves consideration of multiple steps in the environment. We can formally encode the property using an LTLt formula $\varphi = A \rightarrow G$, where $A$ represents the assumptions, and $G$ specifies the guarantees the system must uphold under those assumptions:

\begin{equation*}
    [A] \hspace{.5cm} \square [x = \lhd a + \lhd x]  \hspace{3cm} [G] \hspace{.5cm} \square [\bigwedge_{r\in R}[x \in r] \rightarrow \square_{[1,i]} \neg [x\in r]]
\end{equation*}

\noindent where (i) $A$ (left) is an environment assumption that describes the dynamics of the \textbf{scenario}  and allows to compute the next state given the current state and the action. Specifically, the operator $\lhd x$ represents a specific value that the variable x has taken in the previous timestep; and (ii) $G$ (right) are the guarantees by the system, in this case, stating that if $x\in r$ in a given timestep, then $x\notin r$ throughout $i$ timesteps; i.e., the region $r$ will not be re-visited for $i$ timesteps.
\end{example}

\begin{example} \label{ex:non-markov-b} Note that using the operator $\lhd$ is a very natural way to define A. However, this means that $\varphi$ is encoded in a fragment of LTLt that is not decidable, because $\lhd$ introduces unbounded memory to the program, as it allows an uncontrolled infinite nesting of the operator which at the same time may end up in an infinite recursion. 
Thus, realizability checking is not guaranteed to terminate and we need to rewrite $A$ in a fragment of LTL that does not include $\lhd$, specifically:

\begin{align*}
    [\Tilde{A}] \hspace{.5cm} \square [\bigwedge_{r\in R}[(x+a)\in r] \rightarrow \ocircle [x\in R] ],
\end{align*}

\noindent which means that when the result of applying an action to a state falls within the region $r$, the state itself must belong to $r$ in the next time step. This simple reformulation allows us to eliminate the operator $\lhd$, making realizability of the formula decidable while still preserving dynamic prediction in the subset of information relevant to the requirement $G$.
\end{example}

\noindent An essential observation that makes this transformation possible is that we are (i) assuming to have access to a safety-relevant subset of the robot's dynamic, often called \textit{safety-dynamics} in the literature \citep{yang2023safe}; and (ii) we encoded the requirement for a finite number of regions $r \in R$ and a finite horizon $i$. In our example, we can \textit{anticipate} the constraints-relevant components of the next state given the action and the current state of the system. Thus, we can \textit{anticipate} any unsafe action of the agent, and potentially override it (whenever the specification realizable).

\begin{theorem}
\label{thm:anticipation} 
    If $\varphi$ is an LTLt formula where the subset of the next values of the environment variables that appear in  $\varphi$ are isolated and can be fully determined from the current environment and system values, then $\varphi$ can be rewritten without the use of $\lhd$ (proof in Appendix~\ref{app:anticipation-fragment}).
\end{theorem}

\begin{definition}[Anticipation Fragment]
\label{def:method:anticipation-fragment} 
    The class of LTLt formula that can be translated to not using $\lhd$ is called the anticipation fragment of LTLt.
\end{definition}

\begin{corollary}
\label{cor:anticipation} 
    If $\varphi$ belongs to the anticipation fragment, then the realizability of $\varphi$ is decidable.
\end{corollary}



\section{Case Study: Mapless Navigation with Reinforcement Learning} 
\label{sec:caseStudy}
The agent's goal is to navigate through an unknown environment to reach a target position while avoiding collisions with obstacles. 
This problem is particularly challenging from a safety perspective due to the partial observability of the environment. 
Each episode randomizes the agent's starting point, target destination, and obstacle placements. 
The agent can only perceive its immediate environment via a lidar scan, providing scalar distances to nearby obstacles in a finite set of directions. 
In addition to lidar, observations include the agent's current position, orientation, and target location. 
Actions consist of linear and angular velocities of the robot, and training is performed using Proximal Policy Optimization (PPO)~\citep{schulman2017proximal}, which has been shown to be efficient on similar navigation tasks \citep{amir2023verifying}. 
Additional details on the training environment, reward function, and algorithm are provided in Appendix \ref{sec:app:training}.
For evaluation, we analyze the success and collision rates of trajectories executed by the agent. Fig.\ref{fig:settings:training} illustrates the training process over time. Without any shielding, the agent achieves a success rate of 98.1\% along with a complementarily low collision rate of 1.2\%. However, the collision rate does not reach a constant value of zero, highlighting the persistent risk of unsafe behavior during deployment. This observation underscores the critical need for our proposed shielding approach, which aims to enforce robust safety guarantees and mitigate the remaining risks inherent in complex and dynamic environments.

\subsection{Continuous Shield for Safe Navigation}
In this section, we delineate the safety requirements we aim to enforce and how we specify them using LTLt to ensure robust and reliable performance.

\begin{figure}[t]
  \centering

  \begin{minipage}{0.55\textwidth}
    \centering
    \begin{minipage}{0.47\textwidth}
      \centering
      \includegraphics[width=\textwidth]{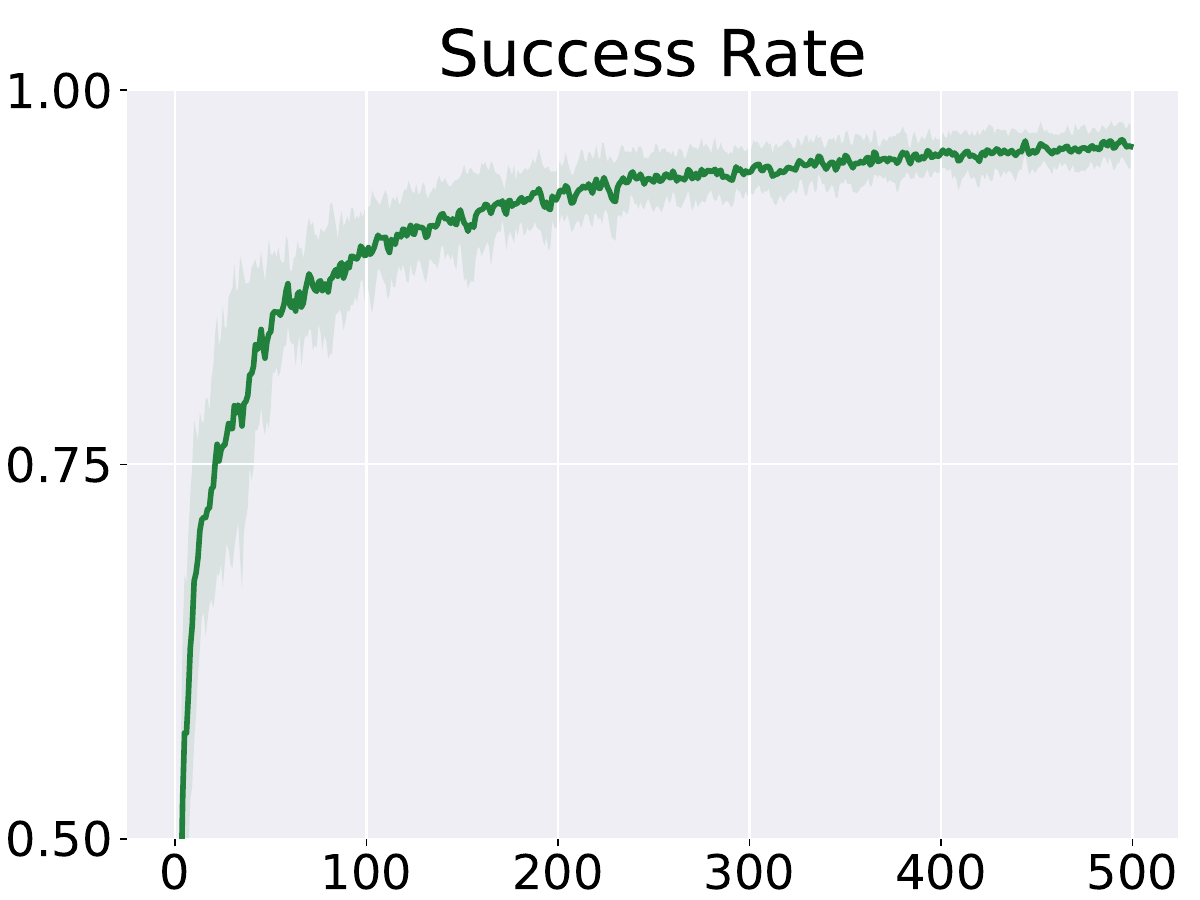}
    \end{minipage}
    \hfill
    \begin{minipage}{0.47\textwidth}
      \centering
      \includegraphics[width=\textwidth]{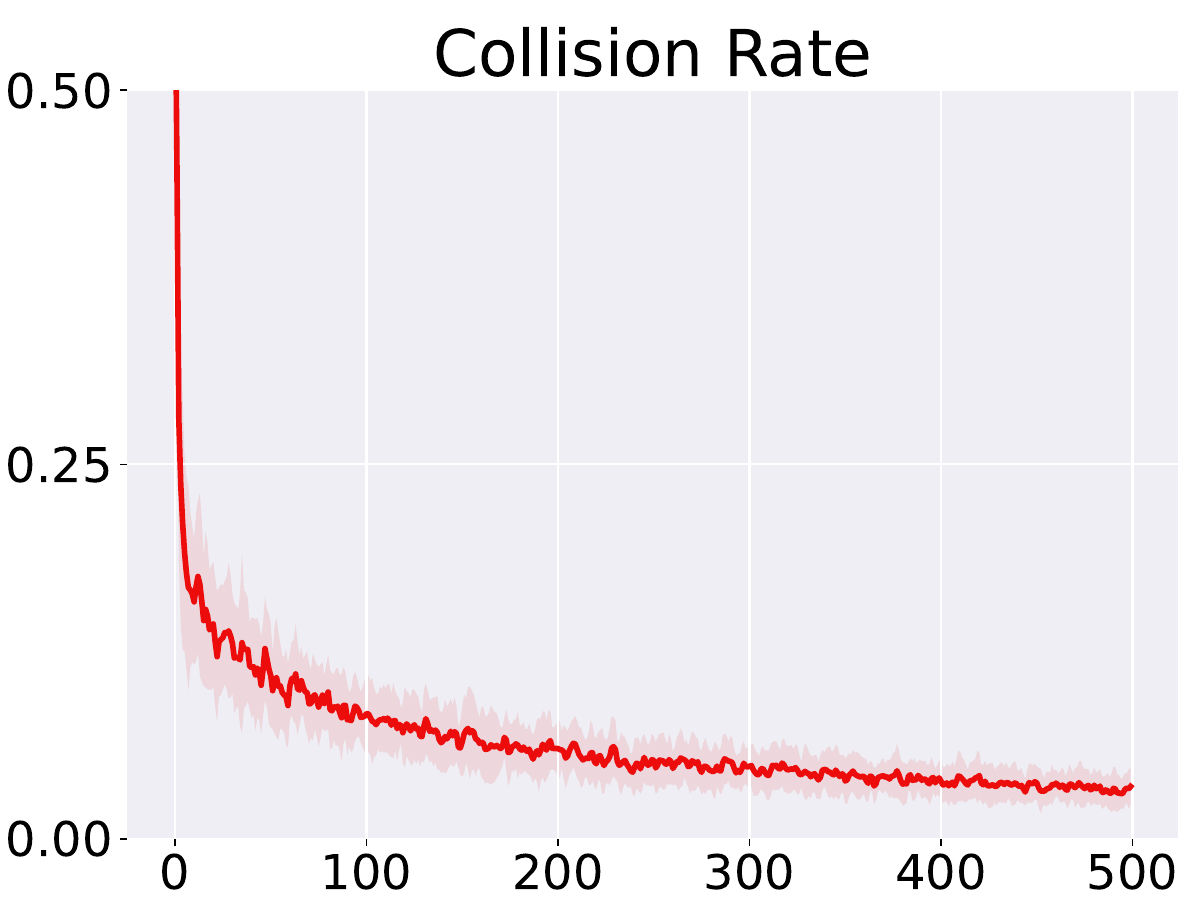}
    \end{minipage}
    
    \caption{Average success rate and collision rate obtained during the DRL training process for 500 episodes (x-axis) without any shield applied (averaged over $5$ different random seeds). 
    }
    \label{fig:settings:training}
  \end{minipage}
  \hfill
  \begin{minipage}{0.4\textwidth}
    \centering
    \includegraphics[width=1\textwidth]{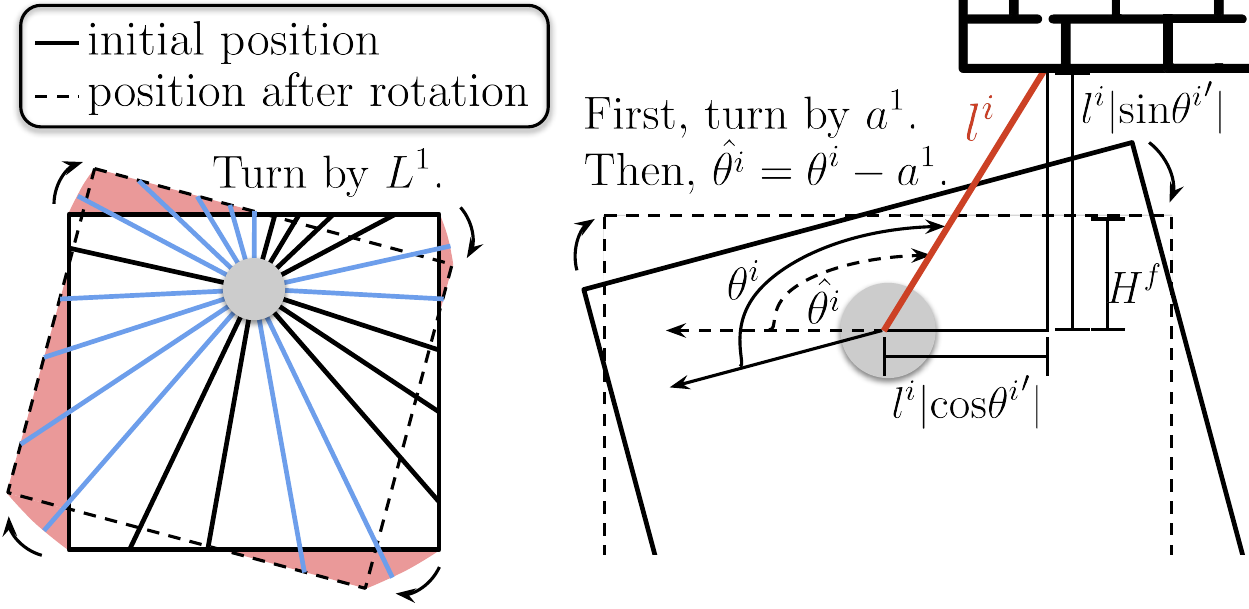}
    \vspace{-14pt}
    \caption{Description of collision avoidance requirements that include the dynamic of the robot as part of the formula.}
    \label{fig:settings:requirement}
  \end{minipage}
\end{figure}

\paragraph{Markovian Requirements for Collision Avoidance.}
In navigation tasks, ensuring collision avoidance is crucial under all circumstances. To achieve this, we encode specifications by leveraging the robot's dynamics, which our framework supports for continuous spaces.
Given an action $a=[a^0,a^1]$, the robot first rotates by $a^1$ and then translates by $a^0$, where $a^0 \in [-L^0, L^0]$ and $a^1 \in [-L^1, L^1]$ and $L^{N}$ is a step size.
Positive $a^0$ indicates forward motion and positive $a^1$ means a right turn. 
As the next pose of the robot is predictable through the dynamics, we can design requirements that strictly avoid collisions in the next timestep. 
Since robot rotation and translation are performed sequentially in the environment dynamics, requirements for each can be designed separately.  
Specifically, the red area in Fig.~\ref{fig:settings:requirement} (left) shows the robot's trajectory when it makes a maximum right turn of $L^1$.
To conservatively prevent collisions from right turns, we prohibit turning right if any lidar value $l^i$ is below a certain threshold $T^{i}$.
These thresholds are precomputed based on the robot's dynamics and maximum step size $L^1$ and they vary for each lidar (highlighted in blue).
Formally, the specification for a right turn is: $\exists i$ s.t. $\, (l^i \leq T^{i}) \rightarrow a^1 \leq 0$.
The same principle applies to left turns.
For translation, we limit the maximum distance that we can translate based on the minimum distance of potential obstacles that any lidars sense in the direction of travel.
Consider the $i^{th}$ lidar with value $l^i$ at an angular position $\theta^i$ relative to the robot's horizontal line, measured clockwise from the left.
Note that $\theta^i$ is known by construction as part of the system.
After the robot rotates by $a^1$, the angular position w.r.t. the new horizontal line becomes $\hat{\theta^i}=\theta^i - a^1$.
If this lidar reveals that the obstacle is in the path of forward translation, the translation $a^0$ cannot exceed the distance to the obstacle; otherwise, a collision will occur.
Formally, the requirement for forward translation w.r.t. the $i^{th}$ lidar is: $l^i|\textrm{cos}{\theta^i}'| \leq \frac{W}{2} \rightarrow a^0 + H^f < l^i|\textrm{sin}{\theta^i}'|$, where $W$ is the robot's width and $H^f$ is the lidar's forward offset.
For forward translation, this requirement must hold for all lidars in front of the robot. Similar requirements apply for backward translation, with different signs, offsets, and considering lidars at the back.

\paragraph{Non-Markovian Requirements for Loop Avoidance.}
While safety remains paramount in autonomous systems, ensuring collision avoidance, etc. alone can lead to overly conservative behavior, often characterized by repetitive or looping actions.
The agent then can be particularly inefficient, wasting time and energy.
To address this issue, we introduce a set of important non-Markovian requirements.
These prohibit repeating the same action in a state for a certain time window.
This encourages the agent to explore new actions, potentially helping it escape from being stuck. 
We encode these requirements using a queue constructed from the shield state of length $L^Q$.
Each queue element consists of the robot's pose (x/y position and rotation) and actions, represented as $(x,y,r,a^0,a^1)$. 
To effectively check for repeated tuples of continuous values, we quantize the pose and action space into grids of $G^P$ and $G^A$ cells, respectively. 
In this way, the shield does not allow an action $(a^0, a^1)$ in a state $(x, y, r)$ if any of the previous tuples in the queue falls into the same cell.
As these requirements take stateful memory as an additional input, they are non-Markovian requirements and however, don't affect the \realizability{} check in our approach.

\subsection{Realizability and Shielding Implementation}

\begin{wraptable}{r}{0.45\linewidth}
\vspace{-15pt}
    \centering
    \begin{tabular}{lc|ccc}
    \toprule
          &  & \multicolumn{3}{c}{\textbf{$G^A$}} \\ 
          &  & \multicolumn{1}{c}{3} & \multicolumn{1}{c}{5} & \multicolumn{1}{c}{30} \\
    \midrule
       \multirow{3}{*}{$L^Q$} & 1   & \cellcolor[HTML]{FF9999}31 & \cellcolor[HTML]{99FF99}0 & \cellcolor[HTML]{99FF99}0 \\
        & 13 & \cellcolor[HTML]{FF9999}288 & \cellcolor[HTML]{FF9999}6 & \cellcolor[HTML]{99FF99}0 \\
        & 100 & \cellcolor[HTML]{FF9999}336 & \cellcolor[HTML]{FF9999}30 & \cellcolor[HTML]{FF9999}0 \\
        \bottomrule
    \end{tabular}
\caption{Number of episodes where the shield returns \texttt{unsat} out of 500 tests, with realizable configurations in green and others in red. \realizability{} checking identifies unsatisfiable specifications that empirical evaluation might overlook. 
}
\label{tab:results:unsat}
\end{wraptable}

We specify the above requirements in LTLt and provide them to the offline \realizability{} check process to guarantee that these requirements are always satisfiable. Our \realizability{} check process follows the implementation established in \cite{rodriguez2023boolean,rodriguez2024adaptive,rodriguez2024realizability,rodriguez2024predictable}. 
After \realizability{} for our requirements is guaranteed, at test-time, we deploy our agent with an online shield to provide safe alternative actions when the agent suggests an action that violates our constraints. 
The overall algorithm for deploying the online shield is presented in Appendix~\ref{sec:app:algorithm}.

\section{Experimental Results}

\paragraph{Realizability Check.}
A crucial aspect of our shield is that it guarantees a safe action in \textit{any} state through a \realizability{} check. Ensuring \realizability{} by hand is challenging with complex requirements like ours, and statistics-driven safety checks are also unreliable. In Table~\ref{tab:results:unsat}, we evaluate a shielded agent for 500 episodes with multiple different queue lengths ($L^Q$) and action grid sizes ($G^A$) without performing a \realizability{} check beforehand. We then tally the number of times we encountered a \texttt{unsat} output from the shield due to no safe action being available. Finally, we verify the \realizability{} of each shield specification and highlight the realizable configuration in green. We see that performing a realizability check can warn us about potentially unsatisfiable and thus unsafe specifications even when empirical evaluations do not indicate a problem. For instance, no \texttt{unsat} situations occurred for our shield with evaluating $(L^Q,G^A)=(100,30)$, however, our \realizability{} check reveals that there still are potential situations where $(100,30)$ is not satisfiable. 
For subsequent experiments, we use one of the verified realizable configurations, $(L^Q,G^A) = (30,13)$.

\paragraph{Online Shielding.}
In Table \ref{tab:results:main}, we report the analysis of 5 different RL agents with different capabilities. \textit{Expert A} and \textit{Expert B} are fully trained PPO models.
\textit{Moderate A} and \textit{Moderate B} are checkpoints collected during an intermediate phase of learning and are more prone to making mistakes and failing the task. Finally, \textit{Unsafe} is \textit{Expert A} deployed without access to lidar data, simulating a dangerous agent oriented toward unsafe behavior and collisions. We believe that the addition of unsafe models is particularly valuable for our analysis to show that the shield can be effective with any model and guarantee safety independently of the input policy.

\begin{table*}[t]
    \centering
    \begin{adjustbox}{width=1\textwidth}
    \begin{tabular}{@{}l|cc|cc|cc|cc@{}}
    \toprule
         & \multicolumn{2}{c}{\textbf{No Shield}} & \multicolumn{2}{c}{\textbf{Collision Shield}} & \multicolumn{2}{c}{\textbf{Collision \& Loop Shield}} & \multicolumn{2}{c}{\textbf{Optimizer}} \\ 
          & Success & Collision & Success & Collision & Success & Collision & Success & Collision \\
    \midrule
        \textit{Expert A} & 0.87 $\pm$ 0.05 & 0.03 $\pm$ 0.03 & 0.87 $\pm$ 0.04 & 0.00 $\pm$ 0.00 & 0.90 $\pm$ 0.02 & 0.00 $\pm$ 0.00 & 0.92 $\pm$ 0.02 & 0.00 $\pm$ 0.00  \\
        \textit{Expert B} & 0.88 $\pm$ 0.03 & 0.03 $\pm$ 0.02 & 0.87 $\pm$ 0.03 & 0.00 $\pm$ 0.00 & 0.90 $\pm$ 0.02 & 0.00 $\pm$ 0.00 & 0.90 $\pm$ 0.02 & 0.00 $\pm$ 0.00  \\
        \textit{Moderate A} & 0.77 $\pm$ 0.04 & 0.01 $\pm$ 0.01 & 0.76 $\pm$ 0.04 & 0.00 $\pm$ 0.00 & 0.79 $\pm$ 0.04 & 0.00 $\pm$ 0.00 & 0.80 $\pm$ 0.05 & 0.00 $\pm$ 0.00  \\
        \textit{Moderate B} & 0.85 $\pm$ 0.02 & 0.04 $\pm$ 0.02 & 0.85 $\pm$ 0.02 & 0.00 $\pm$ 0.00 & 0.87 $\pm$ 0.02 & 0.00 $\pm$ 0.00 & 0.87 $\pm$ 0.02 & 0.00 $\pm$ 0.00  \\
        \textit{Unsafe} & 0.22 $\pm$ 0.03 & 0.78 $\pm$ 0.03 & 0.40 $\pm$ 0.05 & 0.00 $\pm$ 0.00 & 0.47 $\pm$ 0.02 & 0.01* $\pm$ 0.01 & 0.69 $\pm$ 0.03 & 0.01* $\pm$ 0.00  \\
    \bottomrule
    \end{tabular}
    \end{adjustbox}
\caption{Comparison of success rate and collision rate with different settings of the shield on the mapless navigation environment.
Mean scores over 5 seeds (100 runs per seed) with standard deviations are presented.
*These collisions arose due to the partial observability in the environment and can be prevented by increasing the number of 1D lidar sensors.
}
\label{tab:results:main}
\end{table*}

\begin{figure}[t]
\begin{minipage}[t]{0.4\linewidth}
    \includegraphics[width=\linewidth]{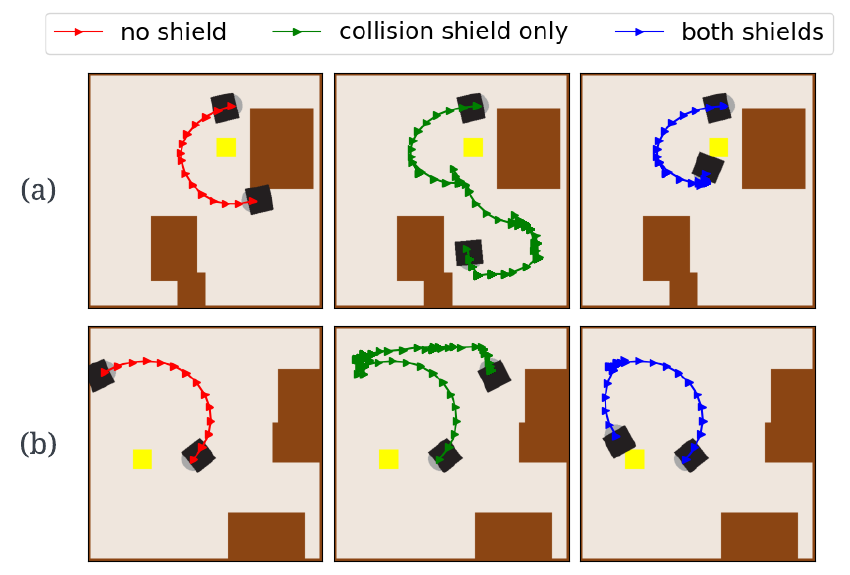}
    \caption{An unshielded agent collides with obstacles and avoids them when a collision shield is applied. By introducing non-Markovian requirements, the agent also avoids oscillations. }
  \label{fig:results:trajectories}
\end{minipage}
\hfill
\begin{minipage}{0.55\textwidth}
    \vspace{-1cm}
    \centering
    \begin{adjustbox}{width=1.0\linewidth}
    \begin{tabular}{@{}l|cc|cc@{}}
    \toprule
         & \multicolumn{2}{c}{\textbf{No Shield}} & \multicolumn{2}{c}{\textbf{Safety Shield}}  \\ 
          & Success & Collision & Success & Collision \\
    \midrule
        \textit{Model A} & 0.56 $\pm$ 0.06 & 0.46 $\pm$ 0.05 & 0.93 $\pm$ 0.02 & 0.00 $\pm$ 0.00 \\
        \textit{Model B} & 0.53 $\pm$ 0.05 & 0.49 $\pm$ 0.07 & 0.95 $\pm$ 0.01 & 0.00 $\pm$ 0.00 \\
        \textit{Model C} & 0.64 $\pm$ 0.03 & 0.36 $\pm$ 0.03 & 0.96 $\pm$ 0.00 & 0.00 $\pm$ 0.00 \\
        \textit{Model D} & 0.66 $\pm$ 0.02 & 0.35 $\pm$ 0.02 & 0.97 $\pm$ 0.01 & 0.00 $\pm$ 0.00 \\
        \bottomrule
    \end{tabular}
    \end{adjustbox}
\captionof{table}{Results on the \texttt{Particle World} environment. The \textit{No Shield} agent is completely blind to the safety requirements and is therefore not trained to avoid collisions. Given the full observability of this environment, it is possible to exploit the dynamics of the system to cover all possible sources of collisions, reducing this number to zero.}
\label{tab:results:particle}
  \end{minipage}
  \vspace{-15pt}
\end{figure}

The first column shows the success rate and collision rate when each policy is executed without any external shielding component. 
Although the success rates for the well-trained agents could be considered satisfactory, all agents made some collisions with an obstacle. The second column shows each policy's performance with the collision avoidance shield added, clearly demonstrating the effectiveness of our shield, which reduces the number of collisions to zero. However, the shield alone leads to overly conservative behavior, resulting in the unsafe trajectories being converted into timeouts with oscillating behavior rather than successful episodes. Crucially, the third column shows the complete version of our shield (i.e., with collision and loop avoidance requirements), showing that the satisfaction of both properties allows the agent to recover from the conservative behavior and increase the success rate while ensuring the safety of the policy. Finally, the last column shows a small additional impact of optimizing the shield's choice of safe action. Since we operate in a continuous space, our shield can employ an optimizer to return a safe action that minimizes distance to the original policy output. In this navigation domain, we minimize $|{\hat{a}^1}-a^1|$ where $a^1$ is the agent's original angular velocity and ${\hat{a}^1}$ is the angular velocity of the shielded action. Fig.~\ref{fig:results:trajectories} illustrates a policy's behavior both with and without a shield employed for collisions and loop.
In the literature, approaches for enhancing safety in deep reinforcement learning provide only empirical results that satisfy requirements in expectation, without formal guarantees \citep{ray2019benchmarking, corsi2022constrained}. While these methods achieve near-optimal safety on average, they fail to fully eliminate violations, underscoring the need for a shield, as discussed in Appendix~\ref{sec:app:related}.

\paragraph{Particle World Experiments}
To further demonstrate the generalizability and robustness of our shielding approach, we also experiment in a multi-agent \texttt{Particle World} environment  \cite{mordatch2017emergence}.
Four agents are tasked with reaching target positions on the opposite side of the map while maintaining a safe distance from one another (a screenshot of this environment can be found in Fig.~\ref{fig:app:environments} of the supplementary materials).
The primary safety requirement in this environment is to ensure that the agents always keep a specified minimum distance between each other, illustrated as circles around the agents.
The results, summarized in Table \ref{tab:results:particle}, demonstrate that our safety shield can be seamlessly applied to this new environment with a continuous state and action space. Notably, our shielding technique in \texttt{Particle World} successfully eliminates all violations of the safety requirements, effectively preventing any unsafe actions. This highlights the versatility and effectiveness of our shielding approach, showcasing its potential for broader applications across various continuous-space environments.

\vspace{-0.8em}
\section{Conclusion}

In this paper, we introduce the concept of \propershield{} and develop a novel approach to creating a shield that can handle continuous action spaces while ensuring realizability. Our results demonstrate that this shielding technique effectively guarantees the safety of our reinforcement learning agent in a navigation problem. A significant achievement of our work is a theoretical and empirical integration of non-Markovian requirements into the shielding process. This integration helps mitigate overly conservative behaviors, enabling the agent to recover from loops and repeated actions caused by strict safety specifications, maintaining efficient progress toward its goals. Limitations and future directions are discussed in the Appendix~\ref{app:limitation}.


\bibliography{l4dc25}

\clearpage \onecolumn \appendix 
\section{Different Shielding Techniques}
\label{sec:app:shield_comparison}

\begin{table*}[h]
\centering
\begin{adjustbox}{width=1.0\linewidth,center}
\begin{tabular}{|c|c|c|c|c|c|c|c|}
\hline
& \begin{tabular}[c]{@{}c@{}}Validate\\ Action\end{tabular} & \begin{tabular}[c]{@{}c@{}}Return Safe \\ Action\end{tabular} & \begin{tabular}[c]{@{}c@{}}Proper Shield \\ (Def.\ref{def:propershield})\end{tabular} & \begin{tabular}[c]{@{}c@{}}Optimization \\ Criteria\end{tabular} & \begin{tabular}[c]{@{}c@{}}Continuous \\ State Space\end{tabular} & \begin{tabular}[c]{@{}c@{}}Non- \\ Markovian\end{tabular} & Real Time             
\\ \hline
Na\"ive Conditions             & \cmark                                     & \xmark                                         & \xmark                                                                                  & \xmark                                            & \cmark                                             & \cmark & \cmark \\ 
\hline
Solver (e.g., Z3)                             & \cmark                                     & \cmark                                         & \xmark                                                                                  & \cmark                                            & \cmark                                             & \cmark & \xmark \\
\hline
\citet{alshiekh2018safe}         & \cmark                                     & \cmark                                         & \cmark                                                                                  & \xmark                                            & \xmark                                             & \xmark & \xmark \\ 
\hline
\citet{rodriguez2023boolean}         & \xmark                                     & \cmark                                         & \xmark                                                                                  & \xmark                                            & \cmark                                            & \xmark  & \xmark \\ 
\hline
This Work & \cmark                                     & \cmark                                         & \cmark                                                                                  & \cmark                                            & \cmark    & \cmark                                         & \xmark \\ \hline
\end{tabular}
\end{adjustbox}
\caption{Comparison of different shielding techniques, highlighting the contribution of our method compared to existing works. 
A na\"ive approach based on a sequence of conditions (e.g., if/else statements) can effectively constrain the action space and prevent undesired decisions, but it cannot return a safe action other than a default one (which is not always available).
A solver-based approach (e.g., Z3) can provide safe alternatives to unsafe actions, but it cannot guarantee the crucial \realizability{} property. This problem has been addressed in the work of \citet{alshiekh2018safe} but not in the case of continuous state and action spaces. Among all these approaches, our algorithm is the only one that can satisfy all these requirements. The last column highlights the remaining open problem of the computational time required to provide a safe alternative action.}
\label{tab:introduction:summary}
\vspace{-8pt}
\end{table*}

\section{Training Details}
\label{sec:app:training}
We ran our experiments on a distributed cluster with 255 CPUs and 1T RAM. Each individual training loop was performed on 2 CPUs and 6GB of RAM, for a total wall time of approximately 15 hours. We perform the training with a customized implementation of the Proximal Policy Optimization algorithm \cite{schulman2017proximal} from the \textit{CleanRL} repository\footnote{\url{https://github.com/vwxyzjn/cleanrl}}. Following is the complete list of hyperparameters and a detailed explanation of the state and action spaces.

\begin{figure}[h]
\centering
\begin{minipage}[t]{0.3\textwidth}
    \centering
    \includegraphics[width=\linewidth]{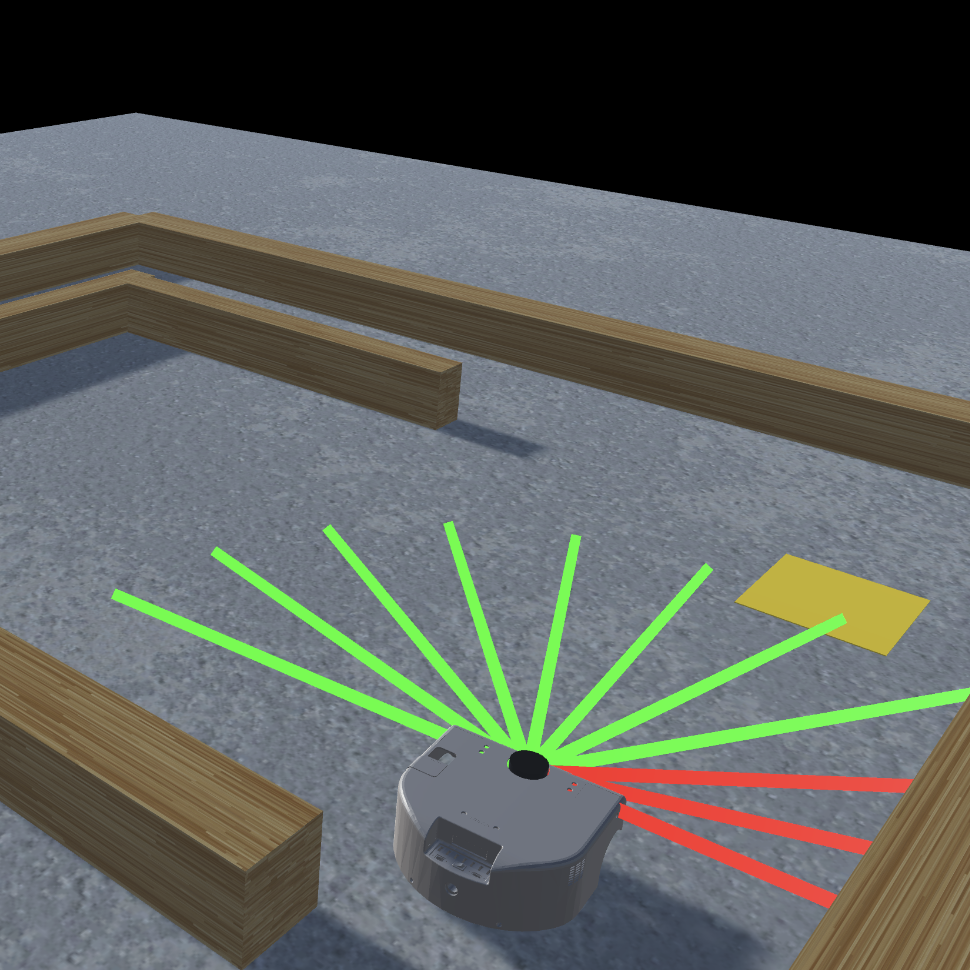}
\end{minipage}
\hspace{1cm}
\begin{minipage}[t]{0.3\textwidth}
    \centering
    \includegraphics[width=\linewidth]{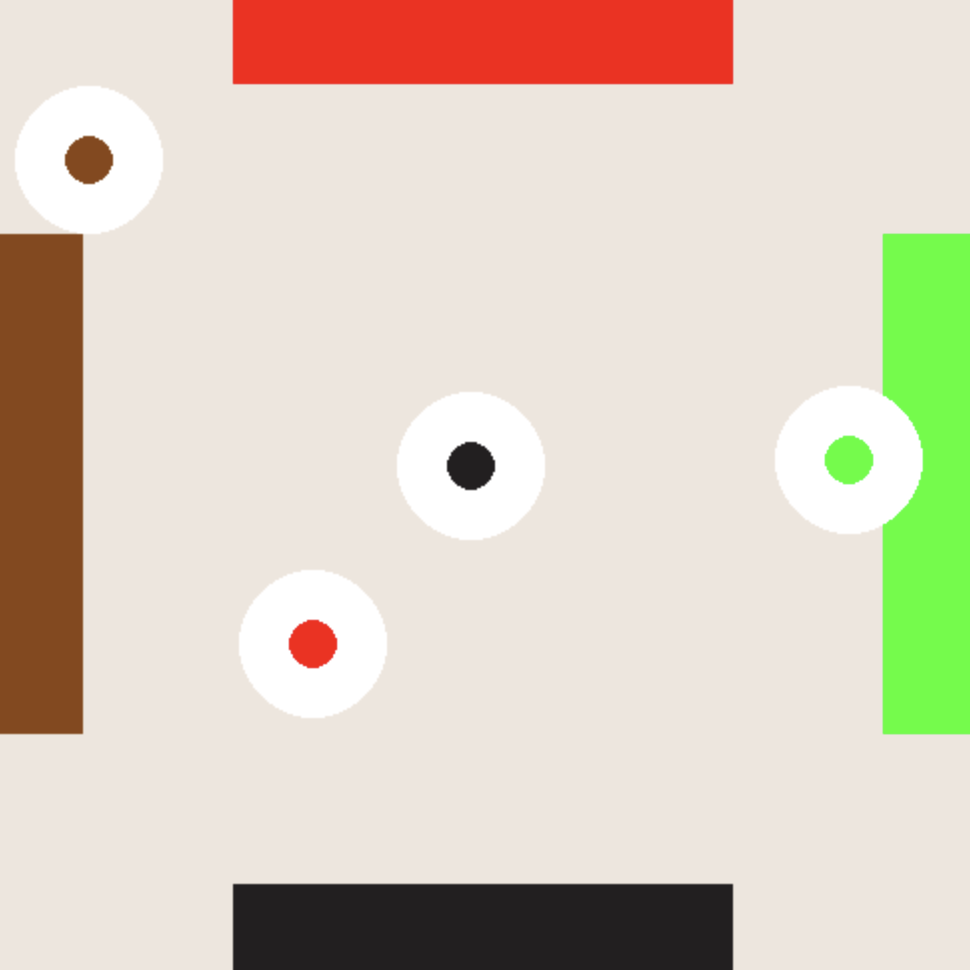}
\end{minipage}
\caption{The \texttt{Mapless Navigation} environment (Left) and the \texttt{Particle World} environment (Right).}
\label{fig:app:environments}
\end{figure}

\paragraph{Reward Function}
For the \texttt{Mapless Navigation} environment, we designed a reward function that incentivize the agent to reach its target position by avoiding the obstacles and additionally provides a penalty to encourage the robot to reach the target with the minimum possible number of steps; more formally:
\[
R(s, a) = \begin{cases}
1 & \text{agent reaches the target} \\
-1 & \text{agent collides with an obstacle} \\
-0.01 & \text{at each timestep otherwise}
\end{cases}
\]

\paragraph{General Parameters}
\begin{itemize}
    \item \textit{training episodes}: 500
    \item \textit{number of hidden layers}: 2
    \item \textit{size of hidden layers}: 32
    \item \textit{activation function}: ReLU
    \item \textit{parallel environments}: 1
    \item \textit{gamma ($\gamma$)}: 0.99
    \item \textit{learning rate}: $0.0015$
\end{itemize}

\paragraph{PPO Parameters}
For the advantage estimation and the critic update, we rely on the Generalized Advantage Estimation (GAE) strategy. The update rule follows the guidelines of the \textit{OpenAI's Spinning Up} documentation\footnote{\url{https://spinningup.openai.com/en/latest/}}. Following is a complete list of the hyperparameters for our training:

\begin{itemize}
    \item \textit{memory limit}: None
    \item \textit{update frequency}: 4096 steps
    \item \textit{trajectory reduction strategy}: sum
    \item \textit{actor epochs}: 30
    \item \textit{actor batch numbers}: 32
    \item \textit{critic epochs}: 30
    \item \textit{critic batch numbers}: 32
    \item \textit{critic network size}: 2x256
    \item \textit{PPO clip}: 0.2
    \item \textit{GAE lambda}: 0.8
    \item \textit{target kl-divergence}: 0.02
    \item \textit{max gradient normal}: 0.5
    \item \textit{entropy coefficent}: 0.001
    \item \textit{learning rate annealing}: yes
\end{itemize}

\paragraph{Random Seeds for Reproducibility:} To obtain the best performing models for our analysis, we applied the following random seeds [$2$, $12$, $18$, $9$, $5$, $1$, $11$] to the following Python modules \textit{Random}, \textit{NumPy}, and \textit{PyTorch}.

\paragraph{State and Action Spaces:} The \texttt{Mapless Navigation} environment follows the structure proposed in \citet{corsi2024verification}, more in details:

\begin{itemize}
    \item The \textit{state space} constitutes of $30$ continuous values: the first $23$ variables represent lidar sensor readings, that indicate the distance from an obstacle in a given direction (from left to right, with a step of $\approx15^{\circ}$). The following $4$ variables represent the (x, y) position of respectively agent and target position. An additional variable indicates the orientation of the agent (i.e., compass). The last $3$ observations are the relative position of the target with respect to the agent in polar coordinates (i.e., heading and distance). All these values are normalized in the interval $[0, 1]$ and can take on any continuous values within this interval.
    
    \item The \textit{action space} consists of $2$ continuous variables: the first one indicates the linear velocity (i.e., the speed of the robot), and the second one provides the angular velocity (i.e., a single value indicating the rotation). These two actions can be executed simultaneously, providing the agent with richer movement options.
\end{itemize}
\section{Algorithm and Implementation Details}
\label{sec:app:algorithm}

In this section, we describe how the shield is deployed in the online process.
Alg.~\ref{alg:test_time_rl} provides a pseudocode of the shielded reinforcement learning loop equipped with the optimizer.
We define general rules for requirements $Reqs$ which is instantiated as $reqs$ for given state $s$ and queue $Q$ for each time step.
Note that these $Reqs$ are verified realizable in the offline process.
And, we have two shield components, $Sol$ and $Opt$, which return an action that respects the $reqs$.
$Opt$ is more advanced as it can bake in additional criteria to optimize, which is possible as our shield is capable of handling continuous spaces.
For example, one effective criterion can be minimizing the distance to the original action $a$.
By doing so, the returned safe action will be closely aligned with the original decision. 
On the other hand, $Sol$ will return \textit{any} safe action that satisfies the specifications which might not be optimal in terms of achieving a goal.
For each time step, we first select an action from the policy $\pi$. If this action satisfies the requirements $reqs$, we perform that action as it is verified safe.
Otherwise, we run the optimizer with a time limit $T_{Opt}$. If the optimizer fails to compute an action in time, we fall back to the action from $Sol$.

\begin{algorithm}
\caption{Shielded Reinforcement Learning Loop}
\label{alg:test_time_rl}
\begin{algorithmic}[1]
\footnotesize{
\REQUIRE Trained RL agent $\pi$, environment $env$, episode length $T$, requirements $Reqs$, solver $Sol$, optimizer $Opt$ with timeout $T_{Opt}$, queue length $L_Q$
\STATE Reset the environment: $s \leftarrow env.reset()$
\STATE Initialize total reward $R \leftarrow 0$ and queue $Q \leftarrow \emptyset$ of length $L_Q$
\FOR{each step $t = 1, 2, \ldots, T$ \textbf{and not} $terminated$}
    \STATE Select action $a \leftarrow \pi(s)$
    \STATE $reqs \leftarrow Reqs(s,Q)$  
    \IF{$a$ satisfies $reqs$}
        \STATE $a_{safe} \leftarrow a$
    \ELSE
        \STATE $a_{opt} \leftarrow Opt(reqs, T_{Opt})$
        \IF{$a_{opt}$ is not $None$}
            \STATE $a_{safe} \leftarrow a_{opt}$
        \ELSE
            \STATE $a_{safe} \leftarrow Sol(reqs)$
        \ENDIF
    \ENDIF
    \STATE $(s, r, terminated) \leftarrow env.step(a_{safe})$
    \STATE Update total reward: $R \leftarrow R + r$
    \STATE $Q.\text{enqueue}(s_x,s_y,s_o,a_{safe}[0],a_{safe}[1])$  \COMMENT{Automatically pops oldest element if length exceeds $L_Q$}
\ENDFOR
\RETURN Total reward $R$
}
\end{algorithmic}
\end{algorithm}
\section{Anticipation Fragment of LTLt}
\label{app:anticipation-fragment}

This appendix extends Subsec.~\ref{subsec:nonMarkov}.
In computational logic, a problem is considered decidable if there is an algorithm that can determine the truth of any statement in a finite amount of time. 
However, reactive synthesis for general Linear Temporal Logic modulo theories (LTLt) is not always decidable. 
For example, synthesis for expressions like $\square(y> \lhd x)$ is undecidable \citep{geatti23decidable}. 

In order to overcome this problem, semi-decidable methods can be used \citep{katis2018validity,wonhyuk2022synthesis}, but 
in safety-critical contexts we are interested in sound and complete processes.
Thus, we need to specify in the decidable Non-Cross State (NCS) \citep{geatti23decidable} fragment that does not allow the $\lhd$ operator.
Although NCS may not look expressive enough for shielding continuous reinforcement learning, we showed in Example~\ref{ex:non-markov} that sometimes we can encode realistic dynamics of the environment in 
the predicates of $\varphi$, without the need of such expressive operator. 
We also called \textit{ancitipation fragment} (Definition~\ref{def:method:anticipation-fragment}) to those formulae that can be rewriten in NCS 
and provided Theorem~\ref{thm:anticipation}, whose intuition is that this is possible because the valuation of 
$v$ in the future timestep can be fully determined from the valuation in the previous timestep.
We now proof sketch this.

\begin{proof}
Let $\varphi$ be an LTLt formula. 
Now, let us assume a subformula $\varphi' \in \varphi$, where $\varphi': \square [v = t(\lhd v_1, \lhd v_2...)]$, 
where $v$ is an environment variable, $v_1,v_2...$ are environment or system variables
and $t$ is a computable function whose input is $v_1,v_2...$ (e.g., $v_1+v_2+...$). 
Also, assume that $v$ appears in predicates $P(v)$ that do not involve other variables
(e.g., $v<10$ is valid, but $v<v'$ is not valid). 
Given these conditions, it is easy to see that we can rewrite any subformula in $\varphi$ containing $P(v)$; 
into a subformula which only depends on the valuations of the past.
In particular, we can construct such formula either (1) using the past LTL operator $Y$ (\textit{yesterday}) 
\cite{lichtenstein85glory} to construct $\varphi_{NCS}: (P(v_1,v_2...)) \wedge \varphi''$, where $\varphi''$ is the rest of $\varphi$;
or (2) by using two temporal layers, i.e., $\varphi_{NCS}: P(v_1,v_2...) \rightarrow \ocircle \varphi''$.
Therefore, by applying this rewriting (and the rest of $\varphi$ remaining unchanged), 
any formula containing $\lhd v$ can be transformed into an NCS formula that does not cross state boundaries
and we will eventually finish rewriting.
\end{proof}

In this paper, our non-Markovian property was not to repeat the visited region, for which the assumptions that
Theorem~\ref{thm:anticipation} needs are enough. 

However, what it can happen in the future that we need more expressivity? Say, 
$\square(v=t(\lhd v_1, \lhd v_2...) \wedge v'=t(\lhd v_1', \lhd v_2'...) \wedge (v<v'))$.
We hypothetise that this can also be done (e.g., using the second rewriting method).
However, further studying the exact conditions for a formula to belong to the
anticipation fragment)is out of the scope of this paper.
Last, note that although we rewrite a formula into NCS, the environment is still reactive, because it can be the case that other inputs are free 
and not determined by the past (e.g., consider a sensor that captures the intensity of the wind).
\section{First-Order Theories} \label{sec:firstOrder}


First-order theories \citep{bradley07calculus} are logical frameworks that use first-order logic (FOL), which extends propositional logic by incorporating quantifiers (i.e., $\forall$ and $\exists$) and variables that can refer to elements in a domain (like integers).
A first-order theory $\mathcal{T}$ consists of: (1) a first-order vocabulary used to define terms and predicates, (2) an interpretation that provides meaning to these terms by mapping them to elements in a domain, and (3) an automatic reasoning system that decides if formulas in the theory are valid.
A theory's \textbf{signature} $\Sigma$ consists of constant, function, and predicate symbols, which are the building blocks of formulas, called $\Sigma$-formula, together with logical connectives and quantifiers.
The symbols in $\Sigma$ initially have no meaning and are given an interpretation later.  
A $\Sigma$-formula $\psi$ is satisfiable if there exists an interpretation in $\mathcal{T}$ that makes $\psi$ true.
A theory $\mathcal{T}$ is decidable if there is an algorithm that can always determine, in a finite amount of time, whether any $\Sigma$-formula $\psi$ is satisfiable.

\paragraph{Arithmetic Theories.}
Arithmetic theories are a particular class of first-order theories which deal with numbers and arithmetic operations.
Many of these theories are decidable, though each comes with a different decision procedure and complexity level.
Typically, these problems are handled using SMT solvers \citep{barrett24satisfiability}.
In this work, we focus on continuous domains represented by Nonlinear Real Arithmetic (NRA) \cite{collins74quantifier}, which is the theory of real numbers with operations like addition and multiplication.  
The signature for NRA is $\Sigma_{\mathbb{R}} = \{0,k,+,-,=,>,\cdot\}$.
For instance, the literal $2x^2>1/3$ is a valid NRA statement.

\section{Safe Reinforcement Learning.}
\label{sec:app:related}

\paragraph{Related Work} Safe reinforcement learning aims to develop policies that not only optimize performance but also respect safety constraints, preventing the agent from taking actions that could lead to harmful or undesirable outcomes. In the literature, this safety aspect can be approached from many different perspectives. A first family of approaches attempts to enforce safer behavior as part of the learning process. An example is constrained reinforcement learning, where the MDP is enriched with a cost function to minimize that represents the safety requirements \cite{he2023autocost, achiam2017constrained, liu2020ipo, ray2019benchmarking}. Other algorithms focus on reward shaping approaches \cite{tessler2018reward}, policy transfer \cite{yang2022training}, behavioral monitors \cite{corsi2022constrained, srinivasan2020learning}, or on techniques that can guarantee safe exploration and data collection phases \cite{marvi2021safe}. 
However, these approaches are not designed to provide formal guarantees about the final policy, especially in the context of deep learning where neural networks are known to be vulnerable to specific input configurations even when trained with state-of-the-art algorithms \cite{szegedy2013intriguing}. Other approaches rely on verification methods to provide guarantees about the behavior of the neural network \cite{katz2019marabou, wang2021beta, katz2017reluplex}; however, they suffer from significant scalability problems and are limited in neural network topology \cite{liu2021algorithms}. Moreover, as many recent papers have shown, as the number of requirements grows, it becomes nearly impossible to find models that formally respect the requirements over the entire input space \cite{corsi2024analyzing, amir2023verifying}.

\paragraph{Safe RL Experiments} To further evaluate the necessity of our shielding approach, we conducted additional experiments using a constrained deep reinforcement learning algorithm \citep{ray2019benchmarking}, specifically designed to enhance safety. As reported in Tab. \ref{tab:app:safe-rl}, while this approach aims to improve the safety of learned policies, we observed that it was still unable to completely eliminate collisions. Moreover, the constrained DRL method provided only marginal improvements over the standard PPO policy, which already demonstrated strong performance in our experiments. These findings underscore the limitations of relying solely on policy optimization, even with safety constraints, to address critical safety requirements. Importantly, our shielding mechanism is not a competing approach but a complementary one, as it can be seamlessly applied on top of any policy, to provide an additional layer of safety during deployment.

\begin{table}[h]
\centering
\begin{tabular}{l|l|l}
\toprule
  & Success & Collision \\
\midrule
$\lambda$-PPO A & 0.86 $\pm$ 0.02 & 0.05 $\pm$ 0.03 \\
$\lambda$-PPO B & 0.87 $\pm$ 0.04 & 0.06 $\pm$ 0.04 \\
$\lambda$-PPO C & 0.85 $\pm$ 0.04 & 0.09 $\pm$ 0.06 \\
$\lambda$-PPO D & 0.88 $\pm$ 0.03 & 0.07 $\pm$ 0.06 \\
$\lambda$-PPO E & 0.86 $\pm$ 0.06 & 0.09 $\pm$ 0.05 \\
\bottomrule
\end{tabular}
\caption{Comparison of success rate and collision rate achieved using the \textit{Lagrangian-PPO} algorithm, a constrained reinforcement learning variant of PPO introduced by \citet{ray2019benchmarking}. The reported models represent the best-performing results selected from a pool of $50$ seeds, and averaged over $5$ runs.}
\label{tab:app:safe-rl}
\end{table}

\section{Limitations and Future Work}
\label{app:limitation}

While our definition of a proper shield ensures a response for every state, an exhaustive realizability check may be overly cautious. Some states may be unreachable by a shielded agent and therefore do not require satisfiability, a factor not accounted for in our current analysis. Exploring this aspect and relaxing these constraints represents an important direction for future research.

Another key limitation in deploying a safe agent in the navigation domain lies in generating safety requirements that are both reasonable and realizable. Moving forward, we aim to simplify the process of generating realizable safety requirements, enabling safe RL agents to be more easily deployed across a broader range of domains.  

\end{document}